\documentclass{article}

\usepackage{arxiv}

\usepackage[utf8]{inputenc} 
\usepackage[T1]{fontenc}    
\usepackage{hyperref}       
\usepackage{url}            
\usepackage{booktabs}       
\usepackage{amsfonts}       
\usepackage{nicefrac}       
\usepackage{microtype}      
\usepackage{lipsum}		
\usepackage{graphicx}
\usepackage{natbib}
\usepackage{doi}

\usepackage{xcolor}         

\usepackage{graphicx}
\usepackage{subfigure}
\usepackage{booktabs} 
\usepackage{amsmath}
\usepackage{amssymb}
\usepackage{amsthm}
\usepackage{caption}
\usepackage{hyperref}
\usepackage{wrapfig}

\usepackage{authblk}
\usepackage{booktabs, makecell}

\theoremstyle{thmstyleone}%
\newtheorem{theorem}{Theorem}
\newtheorem{definition}{Definition}

\title{Graph Polynomial Convolution Models for Node Classification of Non-Homophilous Graphs}

\author[1]{Kishan Wimalawarne}
\author[1,2]{Taiji Suzuki}
\affil[1]{Department of Mathematical Informatics,  The University of Tokyo, Tokyo, Japan}
\affil[2]{Center for Advanced Intelligence Project (AIP), RIKEN, Tokyo, Japan}
\date{}
\begin{document}
\maketitle

\begin{abstract}
We investigate efficient learning from higher-order graph convolution and learning directly from adjacency matrices for node classification.  We revisit the scaled graph residual network and remove ReLU activation from residual layers and apply a single weight matrix at each residual layer. We show that the resulting model lead to new graph convolution models as a polynomial of the normalized adjacency matrix, the residual weight matrix, and the residual scaling parameter. Additionally, we propose adaptive learning between directly graph polynomial convolution models and learning directly from the adjacency matrix. Furthermore, we propose fully adaptive models to learn scaling parameters at each residual layer. We show that generalization bounds of proposed methods are bounded as a polynomial of eigenvalue spectrum, scaling parameters, and upper bounds of residual weights.  By theoretical analysis, we argue that the proposed models can obtain improved generalization  bounds by limiting the higher-orders of convolutions and direct learning from the adjacency matrix.  Using a wide set of real-data, we demonstrate that the proposed methods obtain improved accuracy for node-classification of non-homophilous graphs.
\end{abstract}

\section{Introduction}
Graph convolution networks have  become a highly active research area among the  machine learning and deep learning researchers in recent years. Their success  in many widely growing application areas such as  social influence prediction \cite{li-goldwasser-2019-encoding}, relationship modelling \cite{10.1007/978-3-319-93417-4_38}, recommendation systems \cite{3219819.3219890}, and computer vision \cite{zhaoCVPR19semantic}, have made both the academia and the  industry indentify the significance graph convolution networks. 
Despite the many graph convolution networks available, the optimal use of convolution of features and graph structure  for efficient learning is still widely open for research from perspectives of both model design and theoretical understanding. 

One of the challenging problem in graph convolution networks is learning from graphs with various homophily conditions \cite{doi:10.1146/annurev.soc.27.1.415}. Homophily of a graph dictates the way labeled nodes link with other labeled nodes. When nodes have the tendency to link with other nodes with the same label such graphs are known to be homophilous, while graphs with nodes that have more tendency to link with nodes with different labels are 
known as non-homophilous  \cite{doi:10.1146/annurev.soc.27.1.415,chien2021adaptive}. In early research, most graph convolution networks have been developed by evaluating node classification accuracy against the popular benchmark highly homophilous graph datasets such as Cora, Citeseer, and Pubmed \cite{Kipf:2016tc,pmlr-v97-wu19e}. Recently, learning from non-homophilous graphs has gained considerable attention  and new homophily measures and non-homophilous graph datasets have been introduced \cite{zhu2020beyond,lim2021new} imposing new challenges to existing graph convolution networks.

Graph convolution networks that have been developed for non-homophilous graphs are limited. One of the strategy employed in learning from non-homophilous graphs is  feature learning by using convolutions from higher-order neighbors of nodes. MixHop \cite{mix-hop-haija19a} and GPRGNN \cite{chien2021adaptive} employed this approach by using higher-orders of the normalized adjacency matrix with concatenations and the generalised Pagerank, respectively. However, recent studies \cite{zhu2020beyond,lim2021new} using noval non-homophilous graph datasets  have shown that only learning from node features or the adjacency matrix can  provide competitive accuracy for node classification against popular GCN models. 
\begin{wraptable}{l}{0.5\textwidth}
\center
\begin{tabular}{|l|l|l|}
\hline
Dataset & Squirrel &  Film \\ \hline
MLP & 30.10$\pm$2.00 & \textbf{37.38}$\pm$\textbf{0.85} \\
LINK & \textbf{62.69}$\pm$\textbf{1.69} & 25.24$\pm$1.12  \\
GCN & 46.23$\pm$2.26 & 27.61$\pm$0.95 \\
GPRGNN & 51.19$\pm$1.41 & 35.46$\pm$0.92\\
LINKX & 61.81$\pm$1.80 & 36.10$\pm$1.55 \\ \hline 
\end{tabular}
\caption{Learning only from node features (MLP) or the adjacency matrix (LINK) compared to convolution models}\label{table:intro_results}
\end{wraptable} As we show in the Table \ref{table:intro_results}, learning solely on the adjacency matrix using the LINK \cite{LINK} for Squirrel and learning only using the node features by a MLP for Film  have obtained the best node classification accuracy compared to more sophisticated graph convolution models such as GCN or GPRGNN. The recently developed LINKX \cite{lim2021new} is another simple model using MLPs on node features and the adjacency matrix separately prior to concatenation. Though  LINKX has no explicit graph convolution it has recorded competitive node classification accuracies \cite{lim2021new} for non-homophilous graphs outperforming many state of the art graph convolution networks.  These recent developments have made us re-think on the efficient use of node features and graph strcuture for node classification.  

In this paper, we investigate learning models that adaptively learn by applying adequate graph convolution and direct learning from graph data for a given node classfication problem.  In order to improve the graph convolution, we first revisit the scaled residual graph convolution model and apply simplifying graph convolution by removing ReLU activation from residual layers. We show that the resulting model is a graph polynomial convolution model  with higher-orders of the adjacency matrices similar to the generalized Pagerank. However, our model also has higher-orders of weights making it a polynomial of both adjacency matrices and weights allowing nonlinear feature mixing. 
We further propose a hybrid model to combine the  graph polynomial convolution and direct learning from the adjacency matrix using adaptive scaling. Furthermore, we propose fully adaptive learning models that learn  coefficients of residual layers as well as the scaling parameters between graph convolution and direct learning from the adjacency matrix.  
We analyse generalization bounds of the proposed models using transductive Rademacher complexity and show that our models have better generalization due to polynomial structure of graph convolution and direct learning from adjacency matrices. We evaluate our proposed models on node classification using several benchmark real-data sets  and show that our proposed models give improved performances or comparable performances for non-homophilous graphs compared to existing state of the art methods.

\section{Review}
This section contains a brief overview of graph homophily and graph convolution networks  to provide  motivations for our research. 

We start with stating the notations used in this paper. A graph is represented by $G = (V,E)$ with nodes by $v_i \in V, \; i=1,\ldots,N$ and edges by $(v_i,v_j) \in E$. Let $X \in \mathbb{R}^{N \times q}$ represents a feature matrix with $q$ features. Let $Y \in \mathbb{R}^{N \times C}$ represents $C$-labels of the $N$ nodes.  We consider node classification problem where each node $v \in V$ belongs to a class $y_v \in \{0,1,\ldots,C-1\}$. 
The adjacency matrix of $G$ is represented as $A \in \mathbb{R}^{N \times N}$, and the self-loops added adjacency matrix is  $\hat{A} = A + I_{N}$, where  $I_N \in \mathbb{R}^{N \times N}$ is a identity matrix. We denote the diagonal degree matrix of $\hat{A}$ by $\hat{D}_{ij} = \sum_{k} \hat{A}_{ik}\delta_{ij}$, then the normalized adjacency matrix is $\tilde{A} = \hat{D}^{-1/2}\hat{A}\hat{D}^{-1/2}$.

\subsection{Graph Homophily}
An important property of a graph is the  way labeled nodes form links with adjacent nodes, which is known as the homophily of a graph. 
The homophily ratio \cite{doi:10.1146/annurev.soc.27.1.415,chien2021adaptive} (a.k.a edge homophily \cite{lim2021large}) is  the fraction of edges in a graph that connect nodes with the same class label defined as 
\begin{equation*}
\mathcal{H}(G) = \frac{|\{(u,v):(u,v) \in E \land y_u = y_v \}|}{|E|}. \label{eq:edge_homophily}
\end{equation*}
Based on the edge homophily ratio, a graph is called homophilous when $\mathcal{H}(G) \rightarrow 1$ and  non-homophilous when $\mathcal{H}(G) \rightarrow 0$. 

Recently, \cite{lim2021new,lim2021large} have shown that under class-imbalance conditions, when the majority of the nodes belong to a single class label, then these graphs have edge homophily ratio close to 1. To properly reflect homophily under the class-imbalance condition, a new homophily measure \cite{lim2021new} was proposed as  
\begin{equation*}
\hat{\mathcal{H}}(G) = \frac{1}{C-1} \sum_{k=0}^{C-1}\left[h_k - \frac{|C|}{n} \right]_{+}, \label{eq:new_homopliy_index}
\end{equation*}
where $[a]_{+} = \max\{a,0\}$, $C$ is the number of classes, and $h_k$ is the class-wise homophily matrix defined by
\begin{equation*}
h_k = \frac{\sum_{x \in C_x} d_{x}^{(k_x)}}{\sum_{x \in C_x} d_x},
\end{equation*}
where $d_{x}$ is the number of neighbours, $k_x \in \{0,1,\ldots,C-1\}$ is the class label of the node $x$, and $d_{x}^{(k_x)}$ is the number of neighbors with same label as the node $x$. Similar to the edge homophily, $\hat{\mathcal{H}}(G) \in [0,1]$, and a graph is recognised as non-homophilous when $\hat{\mathcal{H}}(G) \rightarrow 0$.

\subsection{Graph Convolutional Models}
The most basic model to apply graph convolution is the initial  model proposed by \cite{Kipf:2016tc}, which is often referred to as the Vanilla GCN model. This model multiplies node features $X \in \mathrm{R}^{N \times q}$ by the normalized adjacency matrix $\tilde{A} \in \mathrm{R}^{N \times N}$ and apply ReLU activation at each layer. A 2-layer model is  
\begin{equation}
Y = \mathrm{softmax}\big(\tilde{A} \mathrm{ReLU}(\tilde{A}XW_{0})W_{1}\big), \label{eq:gcn}
\end{equation}
where $W_0 \in \mathbb{R}^{q \times h}$ and $W_1 \in \mathbb{R}^{h \times C}$ are learning weights with $h$ hidden units.
\cite{pmlr-v97-wu19e} proposed the simplifying graph convolution (SGC) model by removing the $\mathrm{ReLU}()$ activation from from the Vanilla GCN model \eqref{eq:gcn} as
\begin{equation*}
Y = \mathrm{softmax}\big(\tilde{A}^2XW_{0}W_{1}\big) = \mathrm{softmax}\big(\tilde{A}^2XW\big), \label{eq:sgc}
\end{equation*}
where $W := W_0W_1 \in \mathbb{R}^{q \times C}$.
These models have obtained reasonable node classification accuracy  for homophilous graphs \cite{chien2021adaptive}, however, their accuracy with non-homophilous graphs  have high variance \cite{lim2021large,lim2021new,chien2021adaptive}.

One of the method researchers have successfully employed  when learning from non-homophilous graphs is  convolution of features of a node using features from multiple hops from that node \cite{zhu2020beyond,mix-hop-haija19a}. An efficient method to achieve  such convolution is by higher-orders of the normalized adjacency matrix as used by GPRGNN \cite{chien2021adaptive}. It replaces the normalized adjacency matrix by the Generalized PageRank (GPR) defined as 
\begin{equation}
GPR(\gamma) = \sum_{k=0}^{L-1} \gamma_{k} \tilde{A}^{k}, \label{eq:GPR}
\end{equation}
where $\gamma \in \mathbb{R}^{L}$ are GPR coefficients that are either learned or predefined. 
In GPRGNN, node features are first taken as input to a neural network and then its output is applied with convolution by GPR as   
\begin{equation}
Y = \mathrm{softmax}(Z),\; Z = \sum_{k=0}^{L-1} \gamma_{k} H^{(k)},\;  H^{(k)} = \tilde{A}H^{(k-1)}, \;   H_{i:}^{(0)} = f_{\theta}(X_{i:}).\label{eq:GPR-GNN}   
\end{equation}
GPRGNN has shown be a versatile methods to learn both from homophilous and non-homophilous graphs.

A recent study on node classification of large-scale non-homophilous graphs \cite{lim2021large} has revisited LINK \cite{LINK}, which only use the adjacency matrix to classify nodes without node features or convolution. Simply, LINK \cite{LINK} learns by only using the adjacency matrix as 
\begin{equation*}
Y = \mathrm{softmax}(AW), \label{eq:LINK}
\end{equation*}
where $W \in \mathbb{R}^{n \times C}$ is a weight matrix.
Despite its simplicity, LINK has obtained competitive performances for some non-homophilous graphs compared to well known graph convolution models. LINKX \cite{lim2021large}, an extension of LINK has been proposed  by having two multilinear networks to learn from the adjacency matrix and node features separately and combining their outputs with another multilinear network ($\mathrm{MLP}$) as
\begin{align*}
h_{A} &= \mathrm{MLP}(A) \in \mathbb{R}^{n \times h}   \;\; h_{X} = \mathrm{MLP}(X) \in \mathbb{R}^{n \times h} \nonumber\\
Y &= \mathrm{MLP}( W\mathrm{concat}(h_{A},h_{X}) + h_{A} + h_{X}), 
\end{align*}
where  $\mathrm{concat}()$ is concatenation function, and $W \in \mathbb{R}^{n \times 2h}$. Despite its simplicity LINKX has gained considerable accuracy for non-homophilous graphs  outperforming many graph convolution models \cite{lim2021large}. However, a limitation of both LINK and LINKX is the lack graph convolution, which could make them deprived of efficient feature learning as used by  graph convolution models.

\section{Proposed Method}
In this section, we investigate methods to learn from graphs by optimally using graph convolution and  graph  data (adjacency matrix) to overcome limitations of existing methods.
 
We start by revisiting the residual graph convolution models with $L$ residual layers with scaling by a predefined parameter $\gamma$  as
\begin{align}
X_{0} &= \mathrm{ReLU}(XW_{0}) \nonumber \\
X_{i} &= X_{i-1} + \gamma  \mathrm{ReLU}(\bar{A}X_{i-1} W_{i}), \; i = 1,\ldots,L, \nonumber\\
Y &= \mathrm{softmax}(X_{L}W_{L+1}), \label{eq:residual_scaled}
\end{align}
where $W_0 \in \mathbb{R}^{N \times h}$, $W_i \in \mathbb{R}^{h \times h}$, and $W_{L+1} \in \mathbb{R}^{h \times C}$. We want to remind the reader that applying an appropriate $\gamma$ to \eqref{eq:residual_scaled} leads to a Euler descritization of a graph ordinary differential equation equivalent to GODE \cite{poli2019graph}.

\subsection{Graph Polynomial Convolution Network}
We propose several extensions  the general scaled residual network in \eqref{eq:residual_scaled}. For the basic setting,  we first propose to apply a $T$-layered multilinear network to  the node features $X$. Next, for each scaled residual layers we propose to use only a single weight  $W_1 = W_2 = \cdots = W_L = W_T \mathbb{R}^{h \times h}$. Further, we propose to remove ReLU activation to make \eqref{eq:residual_scaled} to have simplifying graph convolution \cite{pmlr-v97-wu19e}. 
The resulting residual graph convolution model is   
\begin{align}
X_{0} &= \mathrm{ReLU}(XW_{0}) & X_{T+1} &= X_{T} + \gamma \bar{A} X_{T} W_{T}\nonumber \\
X_{1} &= \mathrm{ReLU}(X_{0}W_{1}) & X_{T+2} &= X_{T+1} + \gamma \bar{A} X_{T+1} W_{T}\nonumber \\
&\vdots \nonumber & &\vdots \nonumber\\
X_{T-1} &= \mathrm{ReLU}(X_{T-2}W_{T-1}) &  X_{T+L} &= X_{T+L-1} + \gamma \bar{A} X_{T+L-1} W_{T} \nonumber\\
&\underbrace{\phantom{sssssssssssssss}} \nonumber & &\underbrace{\phantom{ssssssssssssssssssss}} \nonumber\\
&\mathrm{initial\;MLP\;layers} \nonumber & &\mathrm{scaled\;residual\;layers} \nonumber\\
&  & Y = \mathrm{softmax}(X_{T+L}W_{T+1})  &, \label{eq:GPCN} 
\end{align}
where $W_{T+1} \in \mathbb{R}^{h \times C}$.

With simple algebraic operations, \eqref{eq:GPCN} simplifies to the following model   
\begin{align}
X_{0} &= \mathrm{ReLU}(XW_{0}) \nonumber \\
X_{1} &= \mathrm{ReLU}(X_{0}W_{1}) \nonumber \\
&\vdots \nonumber \\
X_{T} &= \mathrm{ReLU}(X_{T-1}W_{T-1}) \nonumber \\
Y = \mathrm{softmax}\bigg( \bigg(X_{T} + \sum_{k=1}^{L-1} L \gamma^{k} &\bar{A}^{k} X_{T} W_T^{k} 
+ \gamma^{L} \bar{A}^{L} X_{T} W_T^{L} \bigg)W_{T+1} \bigg).\label{eq:GPCN_poly}
\end{align}
We name the above models represented by both \eqref{eq:GPCN} and \eqref{eq:GPCN_poly} \textit{Graph Polynomial Convolution Network (GPCN)}.

It is easy to see that \eqref{eq:GPCN_poly} has some similarity to GPR \eqref{eq:GPR} with the sum of higher-orders of the normalized adjacency matrix. It is necessary to identify the main distinctive feature of \eqref{eq:GPCN_poly}  having higher-order of weights different from GPR \eqref{eq:GPR}. The higher-orders of the weight $W_T$ may allow nonlinear mixing of weight parameters at each higher-order convolution compared to linear weight summation as in \eqref{eq:GPR-GNN}. Further, as we demonstrate later with theoretical analysis, higher-order weights allow the model to learn from smaller number of convolution avoiding oversmoothing. We propose to tune both $T$ and $L$ as hyperparameters. 

\subsection{Hybrid Model with Graph Topology}
As we have indicated in the introduction some datasets can give an optimal performance by only learning from graph data with LINK. Hence, we propose to include an exclusive learning component from the adjacency matrix to the proposed models  \eqref{eq:GPCN} and \eqref{eq:GPCN_poly}. A simple way to achieve this is by adding  direct learning with the normalized adjacency matrix and combining it to the graph convolution with a scaling of  $\mu \in [0,1]$ at the output layers of  \eqref{eq:GPCN} and \eqref{eq:GPCN_poly}. With an additional weight $W_{A} \in \mathbb{R}^{N \times h}$ and learning $\mu$ as a parameter, we define GPCN-LINK as
\begin{align}
X_{0} &= \mathrm{ReLU}(XW_{0})  & X_{T+1} &= X_{T} + \gamma \bar{A} X_{T} W_{T}\nonumber \\
X_{1} &= \mathrm{ReLU}(X_{0}W_{1})  & X_{T+2} &= X_{T+1} + \gamma \bar{A} X_{T+1} W_{T}\nonumber \\
\vdots &\nonumber  & &\vdots \nonumber\\
X_{T-1} &= \mathrm{ReLU}(X_{T-2}W_{T-1}) &   X_{T+L} &= X_{T+L-1} + \gamma \bar{A} X_{T+L-1} W_{T} \nonumber\\
Y & = \mathrm{softmax}((\mu X_{T+L} + (1-\mu) \bar{A} W_{A})W_{T+1} .&&   \label{eq:GPCN-LINK}
\end{align}

Using similar simplification  as in \eqref{eq:GPCN_poly} to \eqref{eq:GPCN-LINK}, we obtain  
\begin{align}
X_{0} &= \mathrm{ReLU}(XW_{0}) \nonumber \\
X_{1} &= \mathrm{ReLU}(X_{0}W_{1}) \nonumber \\
&\vdots \nonumber\\
X_{T} &= \mathrm{ReLU}(X_{T-1}W_{T-1}) \nonumber \\
Y = \mathrm{softmax}\bigg( \mu \bigg(X_{T} + \sum_{k=1}^{L-1} L \gamma^{k} \bar{A}^{k} &X_{T} W_T^{k} 
+ \gamma^{L} \bar{A}^{L} X_{T} W_T^{L} \bigg) + (1-\mu) \bar{A} W_{A} \bigg)W_{T+1} \bigg).\label{eq:GPCN-LINK_poly}
\end{align}

\subsection{Adaptive Models}
Instead of having a fixed scaling parameter $\gamma$ for GPCN and GPCN-LINK, we propose to learn  $\gamma$ adaptively at each order of $k$. We can develop different strategies to learn $\gamma$ adaptively as previously explored in GPRGNN \cite{chien2021adaptive, adagpr}. 
However, for simplicity we employ a simple way to make adaptive coefficients to have learnable parameters $\theta \in \mathbb{R}^{L+1}$ and apply L2-norm regularization. The resulting model is   
\begin{align}
X_{0} &= \mathrm{ReLU}(XW_{0}) \nonumber \\
X_{1} &= \mathrm{ReLU}(X_{0}W_{1}) \nonumber \\
&\vdots \nonumber\\
X_{T} &= \mathrm{ReLU}(X_{T-1}W_{T-1}) \nonumber \\
Y = \mathrm{softmax}\bigg( \bigg(&\theta_0 X_{T} + \sum_{k=1}^{L} \theta_k \bar{A}^{k} X_{T} W_T^{k} 
 \bigg)W_{T+1} \bigg).\label{eq:AGPCN}
\end{align}
We call the above model \textit{Adaptive GPCN} (AGPCN). Further, we propose AGPCN-LINK by making both $\gamma$ and $\mu$ in \eqref{eq:GPCN-LINK_poly} adaptive.

\section{Theoretical Analysis}
We analyze generalization bounds of the proposed models using transductive Redemacher complexity  \cite{10.5555/1641503.1641508,OonoS20} under the semi-supervised node classification setting similar to \cite{adagpr}.

We recall that node feature matrix given by  $X \in \mathbb{R}^{N \times q}$  and consider a 1-class labeled output by $Y \in  \mathbb{R}^{N \times 1}$. Let us consider the  sets $\mathcal{X}$ and $\mathcal{Y}$ such that $X \subseteq\mathcal{X}$, $Y \subseteq\mathcal{Y}$ and $(x_i ,y_i) \in \mathcal{X} \times \mathcal{Y}$. We represent $D_{\mathrm{train}}$ and $D_{\mathrm{test}}$ as the training and test sets, respectively and samples are drawn without replacement for $D_{\mathrm{train}}$ and $D_{\mathrm{test}}$ such that $D_{\mathrm{train}} \cup D_{\mathrm{test}} = V$ and  $D_{\mathrm{train}} \cap D_{\mathrm{test}} = \emptyset$. Further, we denote $M: =\vert D_{\mathrm{train}}\vert$ and $U: =\vert D_{\mathrm{test}}\vert$ and define $Q:=1/M + 1/U$. We also specify $C_0,\ldots,C_{L} \in \mathbb{N}_{+}$ with $C_0 = q$, $C_1 = \cdots =C_{T} = h$ and $C_{T+1} = 1$ to represent the dimensions of hidden layers and the output of proposed models.

We analyze the generalization bound for the model GPCN-LINK \eqref{eq:GPCN-LINK_poly} from which we can derive the generalization bounds for other model. We define the hypothesis class for GPCN-LINK by 
\begin{multline}
\mathcal{F}_{\mu,\gamma} = \Big\{ X,\bar{A} \mapsto 
\mathrm{softmax} \bigg(f^{(3)} \circ (\mu f^{(1)} \circ g^{(T)} \circ \cdots \circ g^{(0)}(X) + (1 - \mu)  f^{(2)}(\bar{A})) \vert \\
 g^{(l)}(Z) = \mathrm{Relu}(ZW^{(l)}),  f^{(1)}(X_T) = X_{T} + \sum_{k=1}^{L-1} L \gamma^{k} \bar{A}^{k} X_{T} W^{(T)k} + \gamma^{L} \bar{A}^{L} X_{T} W^{(T)L} \bigg), \\  
f^{(2)}(\bar{A})) = \bar{A}W^{(A)},\; f^{(3)}(Z) = ZW^{(T+1)},\; \|W_{ \cdot c}^{(l)} \|_{1} \leq  B^{(l)}\; 
 \mathrm{for\; all} \; c \in [C_{l+1}], \|W_{ \cdot c}^{(A)} \|_{1} \leq  B^{(A)}
\Big\}, \label{eq:hypo_GPCN-LINK}
\end{multline}
where use notation $W^{(l)}:=W_l$ for \eqref{eq:GPCN-LINK_poly}, $W^{(l)} \in \mathbb{R}^{C_l \times C_{l+1}} \; l = 0,\ldots,T-1$, $W^{(T+1)} := W_{T+1} \in \mathbb{R}^{C_{T+1} \times 1}$, $W^{(A)} := W_{A} \in \mathbb{R}^{N \times C_{T}}$, and $\sigma : \mathbb{R} \rightarrow \mathbb{R}$ is a $1$-Lipschitz function  such that $\sigma(0) = 0$ (e.g. ReLU with output clipping or a Sigmoid function) with bounded output as $\vert\sigma(\cdot)\vert \leq R$, and $B^{(l)}\; l = 0,\ldots,L$ are constants. 

We consider a predictor $h : \mathcal{X} \rightarrow \mathcal{Y},\;\; h \in \mathcal{F}_{\mu,\gamma}$ and  a loss function $l(\cdot,\cdot)$ (e.g., Sigmoid, Sigmoid cross entropy). Now we define the training training error by $R(h) = \frac{1}{M}\sum_{n \in V_{\mathrm{train}}}l(h(x_{n}),y_{n})$ and test error by $\hat{R}(h) = \frac{1}{U}\sum_{n \in V_{\mathrm{test}}}l(h(x_{n}),y_{n})$. Using the standard approach in  \cite{10.5555/1641503.1641508}, we state the generalization bounds for transductive Rademacher complexity $\mathcal{R}(\mathcal{F}_{\mu,\gamma},p)$ with $p \in [0,0.5]$, $S := \frac{2(M+U)\min(M,U)}{(2(M+U)-1)(2\min(M,U) -1)}$, and probability $1-\delta$  as 
\begin{equation}
R(h) \leq \hat{R}(h) + \mathcal{R}(\mathcal{F}_{\mu,\gamma},p_0) + c_0Q \sqrt{\min(M,U)} +  \sqrt{\frac{SQ}{2}\log{\frac{1}{\delta}}},\label{eq:trans_gen_bound}
\end{equation} 
where
\begin{equation*}
\mathcal{R}(\mathcal{V},p) = Q \mathbb{E}_{\epsilon}\bigg[ \sup_{v \in \mathcal{V}} \langle \epsilon , v \rangle \bigg],
\end{equation*}
where $\epsilon = (\epsilon_1,\ldots,\epsilon_N)$ is a sequence of i.i.d. Rademacher variables with distribution $\mathbb{P}(\epsilon_i=1) = \mathbb{P}(\epsilon_i = -1) =p$ and $\mathbb{P}(\epsilon_i = 0) = 1-2p$ and $c_0$ is a constant. We consider the special case where $p=p_0 = MU/(M+U)^2$ as developed in \cite{OonoS20} to arrive at the desired generalization error bound.  Using the above setting, we state the following two theorems for bounds Rademacher complexity of the GPCN-LINK and AGPCN-LINK.

\begin{theorem}
The Rademacher complexity of the GPCN-LINK is bounded as 
\begin{equation*} 
\begin{split}
Q^{-1}\mathcal{\bar{R}}(\mathcal{\tilde{H}}^{(0)}, p)  
&\leq C'B^{(T+1)}\mu\bigg[ 2^{T}\prod_{l=0}^{T-1}B^{(l)} \sqrt{\frac{2MU}{(M+U)^2}}\bigg(I + \sum_{k=1}^{L-1} B^{(T)k} L \gamma^{k} \sum_{j=1}^{N} |\lambda_j|^k  \\
&\quad +  \gamma^{L} B^{(T)L} \sum_{j=1}^{N} |\lambda_j|^L \bigg) \| X \|_{\mathrm{F}}   + \bigg(\sum_{k=1}^{L-1} B^{(T)k} L \gamma^{k} \sum_{j=1}^{N} |\lambda_j|^k \\
&\quad +  \gamma^{L} B^{(T)L} \sum_{j=1}^{N} |\lambda_j|^L \bigg)D\bigg] + (1-\mu)2^{5/2} \frac{B^{(T+1)}B^{(A)}\sqrt{MU}}{(M+U)} \sum_{i=1}^{N} |\lambda_{i}|.\label{eq:theory_GPCN-LINK}  
\end{split} 
\end{equation*}
where $\lambda_{i}$ is the $i$th largest eigenvalue of $\tilde{A}$, $D=\sqrt{N}R$, and $C^{'}$ is a universal constant.
\end{theorem}

\begin{theorem}
The Rademacher complexity of the AGPCN-LINK is bounded as 
\begin{equation*} 
\begin{split}
\mathcal{R}(\mathcal{F}_{\mu,\gamma},p_0)   
&\leq C'B^{(T+1)}\mu\bigg[2^{T}\prod_{l=0}^{T-1}B^{(l)} \sqrt{\frac{2MU}{(M+U)^2}}\bigg(\theta_{0}I + \sum_{k=1}^{L} B^{(T)k} \theta_{k} \sum_{j=1}^{N} |\lambda_j|^k   \bigg)  \prod_{l=0}^{T-1}B^{(l)} \| X \|_{\mathrm{F}}   \\
& \qquad\qquad + \bigg(\sum_{k=1}^{L} B^{(T)k} \theta_{k} \sum_{j=1}^{N} |\lambda_j|^k \bigg)D\bigg] + (1-\mu)2^{5/2} \frac{B^{(T+1)}B^{(A)}\sqrt{MU}}{(M+U)} \sum_{i=1}^{N} |\lambda_{i}|.\label{eq:theory_AGPCN-LINK}  
\end{split} 
\end{equation*}
where $\lambda_{i}$ is the $i$th largest eigenvalue of $\tilde{A}$, $D=\sqrt{N}R$, and $C^{'}$ is a universal constant.
\end{theorem}

By setting $\mu=1$. we obtain bounds for GPCN and AGPCN from Theorem 1 and Theorem 2, respectively.
Similar to the analysis in \cite{adagpr}, since the normalized adjacency matrix has a eigenvalue spectrum of $1 = \lambda_1 \geq \lambda_2 \geq \cdots\geq \lambda_N \geq -1 $, as  the residual layers ($k$) increases the summation of eigenvalues becomes small. Further, due to the polynomial structure of GPCN and GPCN-LINK, when $\gamma <1$ and $B^{(T)}<1$ as the residual layers increase their higher-powers shrink quickly. Hence, models with small number of residual layers would give a better generalization since the effect of very high orders of convolutions are redundant. When $\gamma >1$ or $B^{(T)}>1$, higher-order convolutions may get more prominent, then the bound can be too large. Moreover, it may result in summation of many fast shrinking powers of eigenvalues which  lead to oversmoothing \cite{DBLP:conf/iclr/OonoS20}. Again, oversmoothing can be avoided by selection of smaller number of residual layers. Furthermore, if $\mu$ is very small then models learn mainly by direct learning from the adjacency matrix with limited contribution from graph convolution, leading to less oversmoothing. Similar arguments apply for AGPCN and AGPCN-LINK where the models need to learn $\theta$ to avoid oversmoothing by using appropriate levels of higher orders  convolutions.

\section{Related Methods}
Convolution by higher-orders of the normalized adjacency matrix has been employed by several graph convolution models such as GPRGNN \cite{chien2021adaptive}, MixHop \cite{mix-hop-haija19a}, AdaGPR \cite{adagpr}, and H2GCN \cite{zhu2020beyond}. All these methods have shown that higher-order convolutions can obtain higher accuracy for non-homophilous graphs compared models that apply convolution by a single adjacency matrix. GPCN and their variations differ from all the previous models since it uses  higher powers of weight matrices that constructs a polynomial structure for graph convolution. Furthermore, GPCN-LINK and AGPCN-LINK also employ learning directly from  the adjacency matrix allowing adaptive decoupled learning from graph convolution and graph data. Hence, our proposed methods are significantly different from existing graph convolution models that uses higher-order convolutions.   

The use of direct learning from the adjacency matrix in GPCN-LINK and APGCN-LINK  was inspired by the LINK and LINKX. However, both LINKX \cite{lim2021large} and LINK \cite{LINK} have no graph convolution on node features. Furthermore, we propose to have adaptive scaling between graph convolution and direct learning from adjacency matrix, which makes our methods considerable different from LINK and LINKX. 



\begin{table*}[t]
\center
\begin{tabular}{lllllll} \hline
{\small Method }& {\small  Chameleon }& {\small  Squirrel }& {\small Film/Actor }& {\small Cornell }& {\small Texas  }& {\small Wisconsin }  \\ \hline
{\small Classes} &  5&  5  &5 & 5 & 5 & 5 \\
{\small Nodes} &  2277&   5201 & 7600 & 183 & 183 & 251\\
{\small Edges} &  36101&   198353 & 29926 & 295 & 309 & 499\\
{\small Features} &  2089&   2325 & 931 & 1703 & 1703 & 1703 \\
{\small Edge Homoph.} &  0.247&   0.215 & 0.22 & 0.301 & 0.057 & 0.21  \\
{\small Homophily} &  0.062&   0.025 & 0.011 & 0.047 & 0.001 & 0.094  \\
 \hline
{\small MLP }& {\small   47.89}$\pm${\scriptsize 2.56}  & {\small   30.10}$\pm${\scriptsize 2.00}   & {\small  \textbf{37.38}}$\pm$\textbf{{\scriptsize 0.85}}  & {\small   \textbf{84.86}}$\pm$\textbf{{\scriptsize 6.85}}  & {\small   79.18}$\pm${\scriptsize 6.28} & {\small   81.17}$\pm${\scriptsize 5.96}  \\ 
{\small GCN }& {\small  61.18}$\pm${\scriptsize 2.74 } & {\small  46.23}$\pm${\scriptsize 2.62  } & {\small  27.61}$\pm${\scriptsize 0.95 } & {\small  58.64}$\pm${\scriptsize 5.54 } & {\small  58.92}$\pm${\scriptsize 5.24}& {\small  48.43}$\pm${\scriptsize 4.88 }  \\
{\small SGC }& {\small  63.61}$\pm${\scriptsize 2.55}& {\small  43.71}$\pm${\scriptsize 1.69 } & {\small  27.42}$\pm${\scriptsize 1.16 } & {\small  57.02}$\pm${\scriptsize 5.85 } & {\small  58.64 }$\pm${\scriptsize 5.92 } & {\small   49.21}$\pm${\scriptsize 3.44 } \\
{\small GPRGNN }& {\small  65.76}$\pm${\scriptsize 1.57}& {\small 51.19}$\pm${\scriptsize 1.41 } & {\small  35.46}$\pm${\scriptsize 0.92 } & {\small   \textbf{82.43}}$\pm${\scriptsize \textbf{7.47} } & {\small  \textbf{87.56}}$\pm${\scriptsize  \textbf{1.41} } & {\small  85.88}$\pm${\scriptsize 4.09 }  \\
{\small H2GCN  } & {\small  57.11}$\pm${\scriptsize 1.58}& {\small 36.42}$\pm${\scriptsize 1.89 } & {\small  35.86}$\pm${\scriptsize 1.03 } & {\small  \textbf{82.16}}$\pm${\scriptsize \textbf{4.80} } & {\small  \textbf{84.86}}$\pm${\scriptsize \textbf{6.77} } & {\small   \textbf{86.67}}$\pm${\scriptsize \textbf{4.69}  }   \\
MixHop   & {\small  60.50}$\pm${\scriptsize 2.53 } & {\small  43.80}$\pm${\scriptsize 1.48 } & {\small  32.22}$\pm${\scriptsize 2.34 } & {\small  73.51}$\pm${\scriptsize 6.34 } & {\small  77.84}$\pm${\scriptsize 7.73 } & {\small   \textbf{85.88}}$\pm${\scriptsize \textbf{4.22}  }   \\
{\small LINK}  & {\small  \textbf{72.01}}$\pm${\scriptsize \textbf{1.37} } & {\small  62.69}$\pm${\scriptsize 1.69 } & {\small  25.24}$\pm${\scriptsize 1.12 } & {\small  58.37}$\pm${\scriptsize 3.86 } & {\small  58.91}$\pm${\scriptsize 4.32 } & {\small   48.03}$\pm${\scriptsize 6.63  }   \\
{\small LINKX }& {\small  68.42}$\pm${\scriptsize 1.38 } & {\small  61.81}$\pm${\scriptsize 1.80 } & {\small  36.10}$\pm${\scriptsize 1.55 } & {\small  77.84}$\pm${\scriptsize 5.81 } & {\small  74.60}$\pm${\scriptsize 8.37 } & {\small   75.49}$\pm${\scriptsize 5.72  }   \\   
{\small \textbf{GPCN}}  & {\small  \textbf{71.40}}$\pm${\scriptsize \textbf{1.56}  } & {\small  \textbf{64.30}}$\pm${\scriptsize \textbf{2.24} } & {\small  \textbf{36.77}}$\pm${\scriptsize \textbf{1.10} } & {\small  81.62}$\pm${\scriptsize 6.70 } & {\small  79.45}$\pm${\scriptsize 6.30 } & {\small  85.68}$\pm${\scriptsize 4.88    }  \\
{\small \textbf{GPCN-LINK} }  & {\small  \textbf{71.34}}$\pm${\scriptsize \textbf{3.82} } & {\small  \textbf{67.22}}$\pm${\scriptsize \textbf{1.37} } & {\small  \textbf{37.16}}$\pm${\scriptsize \textbf{0.63} } & {\small  66.21}$\pm${\scriptsize 6.42 } & {\small  67.29}$\pm${\scriptsize 9.85 } & {\small  65.29}$\pm${\scriptsize 14.19   }  \\
{\small \textbf{AGPCN} } & {\small  67.12}$\pm${\scriptsize \textbf{2.51} } & {\small  58.16}$\pm${\scriptsize \textbf{1.63} } & {\small  36.13}$\pm${\scriptsize 1.04 } & {\small  80.0}$\pm${\scriptsize 6.41 } & {\small  \textbf{80.81}}$\pm${\scriptsize \textbf{5.04} } & {\small  \textbf{86.47}}$\pm${\scriptsize \textbf{4.42}   }  \\
{\small \textbf{AGPCN-LINK}  } & {\small  71.07}$\pm${\scriptsize 1.81 } & {\small  \textbf{65.61}}$\pm${\scriptsize \textbf{2.96} } & {\small  36.46}$\pm${\scriptsize 1.03 } & {\small  64.86}$\pm${\scriptsize 8.28 } & {\small  66.21}$\pm${\scriptsize 9.06 } & {\small  68.23}$\pm${\scriptsize 13.66   }  \\\hline
\end{tabular}
\caption{Node-classification accuracy of non-homophilous datasets from \cite{Pei-20-Geom-GCN}. The best three results are highlighted. }
\label{table:old_heterophilic}
\end{table*}

\section{Experiments}
We conducted node classification experiments to evaluate the proposed methods using non-homophilous graphs. 
First, we consider non-homophilous graphs  Chameleon, Squirrel, Cornell, Texas, and Wisconsin from \cite{Pei-20-Geom-GCN}. We used their original data splittings, which randomly split into training, validation and testing sets with  nodes
from each class with percentages of $60\%$, $20\%$, and $20\%$, respectively. Further, we use directed graphs of these datasets as originally used by \cite{Pei-20-Geom-GCN}. 
Additionally, we experimented with selected datasets from the recently introduced  non-homophilous graphs  from \cite{lim2021new,lim2021large}.  Due to computational limitations, we only used five datasets with nodes less than 50000, which include Penn94 \cite{data-penn94}, twitch-gamer \cite{data-twitch}, deezer-europe \cite{data-deezer-europe}, and yelp-chi \cite{data-yelp-chi}. For these datasets we used the $5$ data splits based on train/validation/test sampling of 0.5/0.25/0.25 as used in \cite{lim2021new,lim2021large}.

We performed hyperparameter tuning for the proposed models with hidden $\in \{64,512\}$, learning parameter $\in \{0.01,0.05\}$, weight decay $\in \{0.0, 0.001,0.00001\}$, initial feature learning layers $\in \{1,2,3,4,5\}$, residual layers ($L$) $\in \{1,2,4,8\}$, $\gamma \in \{2^{8},2^{6},\ldots,2^{0},2^{-2},\ldots,2^{-6}\}$, and dropout $\in \{0,0.3,0.6,0.9\}$. For all our experiments we used NVidia GPU V100-PCIE-16GB environment hosted on Intel Xeon Gold 6136 processor servers. A Pytorch implementation of GPCN is available at \href{https://github.com/kishanwn/GPCN}{https://github.com/kishanwn/GPCN}.  

As baseline methods, we considered MLP, GCN \cite{Kipf:2016tc} ,SGC \cite{pmlr-v97-wu19e}, GPRGNN \cite{chien2021adaptive}, H2GCN \cite{zhu2020beyond}, MixHop \cite{mix-hop-haija19a}, LINK\cite{LINK}, and LINKX \cite{lim2021new}. 
Due to the use of same data splittings we borrowed node classification results for baseline methods from \cite{lim2021new,lim2021large}. Since we could not find results with LINK for datasets from \cite{Pei-20-Geom-GCN} and LINKX results for Deezer-europe, Twitch-DE, and Yelp-chi, we performed their experiments using the same hyperparameter settings specified in \cite{lim2021large}. We provide further experiment with homophilous graphs and ablation studies in Section C and Section D of the appendix.

Table \ref{table:old_heterophilic} shows the accuracy for non-homophilous datasets from \cite{Pei-20-Geom-GCN}. An important observation we want to highlight is the high accuracy obtained by LINK for Chameleon and Squirrel compared to well-known graph convolution models.  MLP  has given the lowest accuracy for these two datasets. This indicates that the adjacency matrix contains important features for learning compared to node features. GPCN and GPCN-LINK have given significant improvements for Squirrel and a comparable accuracy compared to LINK for Chameleon. For Film dataset, MLP has given the best performance indicating the node features are more important than graph information, here again, GPCN and GPCN-LINK have given comparable accuracies to MLP. An interesting observation is that only AGPCN has only given a comparable performance for Wisconsin among the small scale graphs. 

Finally, table \ref{table:new_heterophilic} shows results for the new non-homophilous datasets introduced by \cite{lim2021new,lim2021large}. GPCN-LINK and AGPCN-LINK have given have given significant accuracy for Penn94 and Genius compared to all the baseline methods. GPCN has obtained a comparable performance for Twitch-DE with GPCN. For both Yelp-chi and Deezer-Europe, our proposed methods have obtained comparable performance against GPRGNN, however, they have recorded less accuracy compared to H2GCN and MixHop. We further emphasize that AGPCN provide comparable accuracy with GPRGNN for all datasets in table \ref{table:new_heterophilic}. Furthermore, we want to point out that overall our methods have obtained improved accuracy compared to LINKX and LINK.

\begin{table*}[t]
\center
\begin{tabular}{llllll} \hline
{\small Method   }&  {\small Twitch-DE } &{\small  Penn94 } &{\small  Yelp-Chi  }& {\small Deezer-europe  } & {\small Genius }   \\ \hline
{\small  Classes }& {\small  2 }& {\small  2 }& {\small  2 }&  {\small   2 } & {\small   2}\\
{\small  Nodes} & {\small  9498 }& {\small  41,554 } & {\small  45,954} &{\small   28,281 } &{\small   421,961} \\
{\small  Edges} &{\small  153138 }& {\small 1,362,803 } &{\small  3,846,979} &  {\small 92,752}  & {\small 984,979} \\
{\small  Features }& {\small 2545 }&{\small  5 }& {\small  32 }& {\small  31,241 } &{\small  12 }\\
{\small  Edge Homoph.} &{\small  0.632} &{\small  0.470 }& {\small  0.773} & {\small 0.525 } & {\small 0.618} \\
{\small  Homophily }&{\small  0.146 }&{\small  0.046} &{\small   0.052} &{\small  0.030 }  & {\small 0.090} \\
 \hline
{\small   MLP  } & {\small  69.20}$\pm${\scriptsize 0.62 } & {\small  73.61}$\pm${\scriptsize 0.40 } & {\small  \textbf{87.94}}$\pm${\scriptsize \textbf{0.52} } & {\small  66.55}$\pm${\scriptsize 0.72   } & {\small  86.68}$\pm${\scriptsize 0.09 }  \\
{\small  GCN   }& {\small  \textbf{74.07}}$\pm${\scriptsize \textbf{0.68} } & {\small   82.47}$\pm${\scriptsize  0.27}& {\small  63.62}$\pm${\scriptsize 1.00 } & {\small  62.23}$\pm${\scriptsize  0.53   } & {\small  87.42}$\pm${\scriptsize 0.37 } \\
{\small  SGC } & {\small  72.30 }$\pm${\scriptsize 0.22 } & {\small  66.79}$\pm${\scriptsize 0.27 } & {\small  58.62}$\pm${\scriptsize 0.85 } & {\small  59.73}$\pm${\scriptsize 0.12    } & {\small  82.36}$\pm${\scriptsize 0.37 } \\
{\small GPRGNN  } & {\small  \textbf{73.84}}$\pm${\scriptsize \textbf{0.69} } & {\small  84.59}$\pm${\scriptsize 0.29 } & {\small  86.57}$\pm${\scriptsize 0.89 } & {\small  \textbf{66.90}}$\pm${\scriptsize \textbf{0.50}   } & {\small  90.05}$\pm${\scriptsize 0.31 }  \\
{\small H2GCN  }& {\small  72.67}$\pm${\scriptsize 0.65 } & {\small  (M) } & {\small  \textbf{88.48}}$\pm${\scriptsize \textbf{0.21} } & {\small  \textbf{67.22}}$\pm${\scriptsize \textbf{0.90}} & {\small  (M)}\\
{\small MixHop  } & {\small  73.23}$\pm${\scriptsize 0.99 } & {\small  83.47}$\pm${\scriptsize 0.71 } & {\small  \textbf{87.02}}$\pm${\scriptsize \textbf{0.50} } & {\small  \textbf{67.80}}$\pm${\scriptsize \textbf{0.58}  } & {\small  90.58}$\pm${\scriptsize 0.09 }  \\
{\small LINK  } & {\small  72.42}$\pm${\scriptsize 0.57 } & {\small  80.79}$\pm${\scriptsize 0.03 } & {\small  63.44}$\pm${\scriptsize 1.07 } & {\small   57.71}$\pm${\scriptsize 0.36   } & {\small  72.58}$\pm${\scriptsize 0.14 }   \\
{\small LINKX  } & {\small  72.64}$\pm${\scriptsize 0.89 } & {\small  84.71}$\pm${\scriptsize 0.52 } & {\small  76.84}$\pm${\scriptsize 1.82 } & {\small   66.11}$\pm${\scriptsize 0.70 } & {\small  90.77}$\pm${\scriptsize 0.27 }    \\
{\small \textbf{GPCN}  } & {\small  \textbf{73.79}}$\pm${\scriptsize \textbf{0.81} } & {\small  \textbf{84.93}}$\pm${\scriptsize \textbf{0.32}} & {\small  85.86}$\pm${\scriptsize 0.52 } & {\small   64.21}$\pm${\scriptsize 1.43   } & {\small  90.53}$\pm${\scriptsize 0.05 }   \\
{\small \textbf{GPCN-LINK}  } & {\small   73.51}$\pm${\scriptsize 0.73 } & {\small  \textbf{86.16}}$\pm${\scriptsize \textbf{0.26} } & {\small  86.48}$\pm${\scriptsize 0.63 } & {\small   65.69}$\pm${\scriptsize 0.74     } & {\small  \textbf{90.98}}$\pm${\scriptsize \textbf{0.01} }  \\
{\small \textbf{AGPCN}  }& {\small   73.66}$\pm${\scriptsize 0.65 } & {\small 84.73}$\pm${\scriptsize 0.32} & {\small  85.88}$\pm${\scriptsize 0.83} & {\small   66.39}$\pm${\scriptsize 0.51     } & {\small  \textbf{91.79}}$\pm${\scriptsize \textbf{0.06} }  \\
{\small \textbf{AGPCN-LINK} } & {\small  73.41 }$\pm${\scriptsize 0.52 } & {\small  \textbf{86.28}}$\pm${\scriptsize \textbf{0.20} } & {\small 86.22}$\pm${\scriptsize 0.22} & {\small   66.06}$\pm${\scriptsize 0.37     } & {\small  \textbf{91.74}}$\pm${\scriptsize \textbf{0.19} }   \\\hline
\end{tabular}
\caption{Node-classification accuracy of non-homophilous datasets from \cite{lim2021large,lim2021new}. The best three results are highlighted.} 
\label{table:new_heterophilic}
\end{table*}

\section{Conclusions and Future Research}
We propose a new class of graph convolution models that consists of a polynomial of normalized adjacency matrix and weight matrices. We provided multiple levels of adaptive learning for higher-order graph convolution and direct learning from adjacency matrices. Using theoretical analysis, we demonstrate that our methods are able to obtain better generalization bounds by avoiding unnecessary higher-order convolution and by mixed learning of graph convolution and direct learning from adjacency matrices. 
By experiments with non-homophilous graphs we demonstrated that our proposed methods can obtained improved performance for node classification compared to many state of the art graph convolution models.

A useful possible extension of the GPCN is to develop  generalized polynomial convolution models where residual layers have different weights. Further research in the direction of adaptive learning may help to reduce the considerable number of hyperparameters. 

\section{Societal Impacts and Limitations}
 Node classification is important application in multiple domains. Further, there are many real-world domains with non-homophilous graphs. However, learning on node features and graph topology may lead to violation of privacy of individuals. A potential practical limitation of the proposed is the considerable number of hyperparameters that needs to be tuned.



{\small
\bibliography{ref}
\bibliographystyle{apalike}
}

\appendix
\section{Generalized Polynomial Model}
For completeness, we provide the model of removing ReLU activations from the scaled residual layers having different weight matrices, which leads to the generalized polynomial model. The resulting extention to \eqref{eq:residual_scaled} is
\begin{align*}
X_{0} &= XW_{0} \nonumber \\
X_{1} &= X_{0} + \gamma \bar{A} X_{0} W_{1} \nonumber\\
X_{2} &= X_{1} + \gamma \bar{A} X_{1} W_{2} \nonumber\\
&\vdots \nonumber\\
X_{L} &= X_{L-1} + \gamma \bar{A} X_{L-1} W_{L} \nonumber\\
Y &= \mathrm{sigmoid}(X_{L+1}W_{2}), 
\end{align*}
which leads a generalized polynomial model  of
\begin{multline*}
Y = \mathrm{sigmoid}\bigg( \bigg(XW_{0} + \gamma \bar{A}(XW_{0})(W_1 + \cdots + W_L) + \gamma^2 \bar{A}^{2}(XW_{0})(W_1W_2 + \cdots + W_1W_L)  \\
+ \cdots  \gamma^{k} \bar{A}^{k} (XW_{0}) W_1W_2 \cdots W_L\bigg)W_{L+1} \bigg).
\end{multline*}
It is easy to see that above models reduces to GPCN models if the same weight is applied to each residual layer. We propose the study of the generalized polynomial convolution models as a future research direction.

\section{Proof of Theoretical Results}
We state the transductive Rademacher complexity \cite{10.5555/1641503.1641508} in the following definition. 

\begin{definition}
Given $p \in [0,0.5]$ and $\mathcal{V} \subset \mathbb{R}^{N}$, the transductive Rademacher complexity is defined as
\begin{equation*}
\mathcal{R}(\mathcal{V},p) = Q \mathbb{E}_{\epsilon}\bigg[ \sup_{v \in \mathcal{V}} \langle \epsilon , v \rangle \bigg],
\end{equation*}
where $Q = \frac{1}{M} + \frac{1}{N}$ and $\epsilon = (\epsilon_1,\ldots,\epsilon_N)$ is a sequence of i.i.d. Rademacher variables with distribution $\mathbb{P}(\epsilon_i=1) = \mathbb{P}(\epsilon_i = -1) =p$ and $\mathbb{P}(\epsilon_i = 0) = 1-2p$. 
\end{definition}

We borrow the following symmetric Rademacher complexity from \cite{OonoS20}, which is a variant of the tranductive Rademacher complexity in Definition 1.

\begin{definition}
Given $p \in [0,0.5]$ and $\mathcal{V} \subset \mathbb{R}^{N}$, the symmetric transductive Rademacher complexity is defined as
\begin{equation*}
\mathcal{\bar{R}}(\mathcal{V},p) = Q \mathbb{E}_{\epsilon}\bigg[ \sup_{v \in \mathcal{V}} \vert\langle \epsilon , v \rangle\vert \bigg],
\end{equation*}
where $Q = \frac{1}{M} + \frac{1}{N}$ and $\epsilon = (\epsilon_1,\ldots,\epsilon_N)$ is a sequence of i.i.d. Rademacher variables with distribution $\mathbb{P}(\epsilon_i=1) = \mathbb{P}(\epsilon_i = -1) =p$ and $\mathbb{P}(\epsilon_i = 0) = 1-2p$. 
\end{definition}


\textit{Proof of Theorem 1.}{
From \cite{OonoS20}, it si known that $\mathcal{R}(\mathcal{F},p) \leq \mathcal{\bar{R}}(\mathcal{F},p)$. Hence, we bound  \eqref{eq:trans_gen_bound} using the symmetric Rademacher complexity. We give the bound for a general $p$ but the final bound can be obtained by substituting $p \leftarrow p_0$.

We use the abbreviation for the row $s$ of any matrix $Z$ by $Z_s := Z[s,:]$, columns $c$ by $Z_{\cdot c} := X[:,c]$, and an element by $Z_{sc} := Z[s,c]$.
We decompose the hyperthesis class in \eqref{eq:hypo_GPCN-LINK} into several components as
\begin{gather}
\mathcal{G}^{(0)} = \Bigg\{ \sum_{c=1}^{C_0} X_{\cdot c}w^{(0)}_{c} \vert  \| w^{(0)}_{c} \|_{1} \leq B^{(0)}  \Bigg\}, \nonumber \\
\mathcal{\tilde{G}}^{(0)} = \sigma \circ \mathcal{G}^{(0)}, \nonumber\\
\mathcal{G}^{(l)} = \Bigg\{ \sum_{c=1}^{C_l} Z_{\cdot c}w^{(l)}_{c} \vert  \| w^{(l)}_{c} \|_{1} \leq B^{(l)} , Z \in \mathcal{G}^{(l-1)}  \Bigg\}, \nonumber\\
\mathcal{\tilde{G}}^{(l)} = \sigma \circ \mathcal{G}^{(l)}, \; l = 1,\ldots,T-1, \nonumber\\
\mathcal{H}^{(0)} = \Bigg\{G + \sum_{k=1}^{L-1} \sum_{c=1}^{C_{T}} K \gamma^{k}[\bar{A}^{k}G]_{\cdot c}w_{c}^{(T)k} + \sum_{c=1}^{C_{T}}\gamma^{K}[\bar{A}G]_{\cdot c}w_{c}^{(T)k}  \bigg\vert    G \in \mathcal{\tilde{G}}^{(T)}, \| w^{(T)}_{c} \|_{1} \leq B^{(T)}  \Bigg\}, \nonumber\\
\mathcal{\tilde{H}}^{(0)} = \sigma \circ \mathcal{H}^{(0)} \nonumber\\
\mathcal{H}^{(1)} = \bigg\{ \bar{A}_{\cdot c}w_{c}^{(A)} \bigg\vert   \| w^{(A)}_{c} \|_{1} \leq B^{(A)}  \bigg\}. \label{eq:simplyfied_hypo} 
\end{gather}

Now, let us consider the output layer, then

\begin{equation} 
\begin{split}
Q^{-1}\mathcal{\bar{R}}(\mathcal{\tilde{F}_{\mu,\gamma}}, p)  &= \mathbb{E}_{\boldsymbol{\epsilon}}  \bigg[\sup_{\|w^{(T+1)}_c \|_1 \leq B^{(T+1)}, Z^{(0)} \in \mathcal{H}^{(0)}, Z^{(1)} \in \mathcal{\tilde{H}}^{(1)}  } \bigg\vert \sum_{n=1}^{N} \epsilon_{n} \sum_{c=1}^{C_{T+1}} \bigg( \mu Z^{(0)}_{n c}  + (1-\mu) Z^{(1)}_{n c} \bigg)w^{(T+1)}_{c}  \bigg\vert \bigg]  \\
&= \mathbb{E}_{\boldsymbol{\epsilon}}  \bigg[\sup_{\|w^{(T+1)}_c \|_1 \leq B^{(T+1)}, Z^{(0)} \in \mathcal{H}^{(0)}, Z^{(1)} \in \mathcal{\tilde{H}}^{(1)}  } \bigg\vert \sum_{c=1}^{C_{L+1}} \sum_{n=1}^{N} \epsilon_{n}  \bigg( \mu Z^{(0)}_{n c}  + (1-\mu) Z^{(1)}_{n c} \bigg)w^{(T+1)}_{c}  \bigg\vert \bigg]  \\
&\leq B^{(T+1)}\mathbb{E}_{\boldsymbol{\epsilon}}  \bigg[\sup_{Z^{(0)} \in \mathcal{H}^{(0)}, Z^{(1)} \in \mathcal{\tilde{H}}^{(1)}  } \bigg\vert \sum_{n=1}^{N} \epsilon_{n}  \bigg( \mu Z^{(0)}_{n }  + (1-\mu) Z^{(1)}_{n } \bigg) \bigg\vert \bigg]  \\
&= B^{(T+1)}\mathbb{E}_{\boldsymbol{\epsilon}}  \bigg[\sup_{\|w^{(T)}_c \|_1 \leq B^{(T)},\|w^{(A)}_c \|_1 \leq B^{(A)}, G \in \mathcal{G}^{(T-1)} } \bigg\vert \\ & \qquad\qquad\qquad \sum_{n=1}^{N} \epsilon_{n} \bigg( \mu \bigg(G_{n} + \sum_{k=1}^{L-1} \sum_{c=1}^{C_{L+1}} K \gamma^{k}[\bar{A}^{k}G]_{n c}w_{c}^{(T)k} + \sum_{c=1}^{C_{L+1}}\gamma^{K}[\bar{A}^{K}G]_{n c}w_{c}^{(T)k} \bigg) \\
& \qquad\qquad\qquad\qquad\qquad\qquad\qquad\qquad   
 + (1-\mu) \sum_{c=1}^{C_{L+1}}\bar{A}_{n c}w_{c}^{(A)} \bigg) \bigg\vert \bigg] \\
&\leq B^{(L+1)}\mathbb{E}_{\boldsymbol{\epsilon}}  \bigg[ \mu\sup_{\|w^{(T)}_c \|_1 \leq B^{(T)}, G \in \mathcal{G}^{(T-1)} } \bigg\vert \sum_{n=1}^{N} \epsilon_{n} \bigg(G_{n} + \sum_{k=1}^{L-1} \sum_{c=1}^{C_{L+1}} K \gamma^{k}[\bar{A}^{k}G]_{n c}w_{c}^{(T)k} \\
& \qquad\qquad\qquad\qquad\qquad\qquad\qquad\qquad   
+ \sum_{c=1}^{C_{L+1}}\gamma^{K}[\bar{A}^{K}G]_{n c}w_{c}^{(T)k} \bigg) \bigg\vert \\
& \qquad\qquad\qquad\qquad\qquad + (1-\mu) \sup_{\|w^{(A)}_c \|_1 \leq B^{(A)}, G \in \mathcal{G}^{(T-1)} } \bigg\vert \sum_{n=1}^{N} \epsilon_{n} \sum_{c=1}^{C_{L+1}}\bar{A}_{n c}w_{c}^{(A)} \bigg\vert \bigg] \\
\end{split} \label{eq:reduction_rad-step1}
\end{equation}

\begin{equation} 
\begin{split}
Q^{-1}\mathcal{\bar{R}}(\mathcal{\tilde{F}_{\mu,\gamma}}, p)  &\leq 
 B^{(T+1)}\mathbb{E}_{\boldsymbol{\epsilon}}  \bigg[ \mu\sup_{\|w^{(T)}_c \|_1 \leq B^{(T)}, G \in \mathcal{G}^{(T-1)} } \bigg\vert  \sum_{n=1}^{N} \epsilon_{n} G_{n} \bigg\vert  \\
& \qquad \qquad + \bigg\vert \sum_{c=1}^{C_{L+1}} \sum_{k=1}^{L-1} \sum_{n=1}^{N} \epsilon_{n}  K \gamma^{k}[\bar{A}^{k}G]_{n c}w_{c}^{(T)k} \bigg\vert 
+ \bigg\vert \sum_{c=1}^{C_{T}} \sum_{n=1}^{N} \epsilon_{n}\gamma^{K}[\bar{A}^{K}G]_{n c}w_{c}^{(T)k}  \bigg\vert \\
& \qquad\qquad\qquad\qquad\qquad + (1-\mu) \sup_{\|w^{(A)}_c \|_1 \leq B^{(A)}} \bigg\vert \sum_{c=1}^{C_{L+1}} \bigg( \sum_{n=1}^{N}  \epsilon_{n} \bar{A}_{n c}\bigg)w_{c}^{(A)}  \bigg\vert \bigg] \\
&\leq B^{(T+1)}\mathbb{E}_{\boldsymbol{\epsilon}}  \sup_{G \in \mathcal{G}^{(T-1)} } \bigg[ \bigg\vert  \sum_{n=1}^{N} \epsilon_{n} G_{n} \bigg\vert  + \sum_{k=1}^{K-1} B^{(T)k} K \gamma^{k} \bigg\vert  \sum_{n=1}^{N} \epsilon_{n}  [\bar{A}^{k}G]_{n} \bigg\vert \\
& \qquad\qquad + \gamma^{K}  B^{(T)K}\bigg\vert \sum_{n=1}^{N} \epsilon_{n}[\bar{A}^{K}G]_{n }  \bigg\vert \bigg]     
+   (1-\mu) B^{(T+1)}B^{(A)}\mathbb{E}_{\boldsymbol{\epsilon}}  \bigg\vert  \bigg( \sum_{n=1}^{N}  \epsilon_{n} \bar{A}_{n }\bigg)  \bigg\vert,  
\end{split} \label{eq:reduction_rad-step2}
\end{equation}
where we have used the inequality $\|W^{(T)k}_c \|_1 \leq \|W^{(T)}_c \|_1^k \leq B^{(T)k}$.

Next, we reduce each component with expectation over $\epsilon$ in \eqref{eq:reduction_rad-step2}. Hence, we consider the bound of the term 
\begin{equation}
\Pi = \mathbb{E}_{\boldsymbol{\epsilon}}  \sup_{G \in \mathcal{G}^{(T-1)} } \bigg\vert  \sum_{n=1}^{N} \epsilon_{n}  [\bar{A}^{k}G]_{n} \bigg\vert. \label{eq:bound_term1}
\end{equation}

Let $\epsilon'=(\epsilon'_1,\dots,\epsilon'_N)$ be a random variable that 
is independent to and has the identical distribution as $\epsilon$.
Then, using a similar methods as in \cite{adagpr} we have that 
\begin{align*}
\Pi &= \mathbb{E}_\epsilon\left[\sup_{ G \in \mathcal{G}^{(T-1)} } \left\vert\sum_{n=1}^N\epsilon_n [\tilde{A}^{k} \frac{1}{2p}\mathbb{E}_{\epsilon'}[\epsilon'\epsilon'^\top]G]_n \right\vert\right]~~(\because \mathbb{E}_{\epsilon'}[\epsilon'\epsilon'^\top] = 2p I) \\
& \leq 
\frac{1}{2p}\mathbb{E}_{\epsilon,\epsilon'}\left[\sup_{ G \in \mathcal{G}^{(T-1)} } \left\vert\epsilon^\top \tilde{A}^{k} \epsilon' \epsilon'^\top G \right\vert\right] 
\leq 
\frac{1}{2p}\mathbb{E}_{\epsilon,\epsilon'}\left[\left\vert\epsilon^\top \tilde{A}^{k} \epsilon'  \right\vert \sup_{ G \in \mathcal{G}^{(T-1)} } \left\vert\epsilon'^\top G \right\vert\right] \\
& =
\frac{1}{2p}\mathbb{E}_{\epsilon'}\left[\mathbb{E}_{\epsilon}\left[\left\vert\epsilon^\top \tilde{A}^{k} \epsilon'  \right\vert \right]\sup_{ G \in \mathcal{G}^{(T-1)} } \left\vert\epsilon'^\top G \right\vert\right] \\
& \leq
\frac{1}{2p}\mathbb{E}_{\epsilon'}\left[\sqrt{\mathbb{E}_{\epsilon}\left[\left(\epsilon^\top \tilde{A}^{k} \epsilon'  \right)^2 \right]} \sup_{ G \in \mathcal{G}^{(T-1)} } \left\vert\epsilon'^\top G \right\vert\right] \\
&=
\frac{1}{2p}\mathbb{E}_{\epsilon'}\left[\sqrt{\mathbb{E}_{\epsilon}\left[\epsilon'^\top \tilde{A}^{k} \epsilon \epsilon^\top \tilde{A}^{k}  \epsilon' \right]} \sup_{ G \in \mathcal{G}^{(T-1)} } \left\vert\epsilon'^\top G \right\vert\right]
\\
& =
\mathbb{E}_{\epsilon'}\left[\sqrt{ \epsilon'^\top \tilde{A}^{2k} \epsilon'  } \sup_{ G \in \mathcal{G}^{(T-1)} } \left\vert\epsilon'^\top G \right\vert\right],
\end{align*}
where we used $\mathbb{E}_{\epsilon}[\epsilon\epsilon^\top] = 2p I$ in the last equation. 
Again, as in \cite{adagpr} we use  Hanson-Wright concentration inequality (Theorem 2.5 of \cite{adamczak2015note}), which results in 
$$
\mathbb{P}[|\epsilon'^\top \tilde{A}^{2k} \epsilon' - \mathbb{E}_{\epsilon'}[\epsilon'^\top \tilde{A}^{2k} \epsilon']| \geq c(\sqrt{2p} \|\tilde{A}^{2k}\|_F \sqrt{ t} + \|\tilde{A}^{2k}\| t)] \leq \exp(- t)~~~(t>0),
$$
with a universal constant $c$, where $\|A\|_F = \sqrt{\mathrm{Tr}[AA^\top]}$.
Further, the Talagrand's concentration inequality yields 
\begin{multline}
\mathbb{P} \Bigg[  \left| \sup_{G \in \mathcal{G}^{(T-1)} } \epsilon'^\top G \right|  \geq c'\Bigg( \mathbb{E}_{\epsilon'}\left[\sup_{G \in \mathcal{G}^{(T-1)} } \left| \epsilon'^\top G \right| \right] \\ 
 + \sqrt{ N t \sup_{G \in \mathcal{G}^{(T-1)} }\sum_{n=1}^N G_n^2/N  } +   t  \sup_{G \in \mathcal{G}^{(T-1)} }\|G\|_\infty  \Bigg)  \Bigg]
\leq e^{-t}~~~(t > 0),
\end{multline}
where $c' > 0$ is a universal constant.
Then, by noticing that $\mathbb{E}_{\epsilon'}[\epsilon'^\top A \epsilon'] = 2p \mathrm{Tr}[A]$, these inequalities yield
\begin{align*}
& \mathbb{E}_{\epsilon'}\left[\sqrt{ \epsilon'^\top \tilde{A}^{2k} \epsilon'  } \sup_{G \in \mathcal{G}^{(T-1)} } \left|\epsilon'^\top G \right|\right] \\
\leq &
\int \sqrt{2p \mathrm{Tr}[\tilde{A}^{2k}] + c (\sqrt{2p}\|\tilde{A}^{2k}\|_F \sqrt{t} + \|\tilde{A}^{2k}\|t) } \\
& \qquad\qquad\qquad\qquad\qquad\qquad\qquad\qquad c' ( | \mathbb{E}_{\epsilon'} \sup_{G \in \mathcal{G}^{(T-1)} } \epsilon'^\top G |  + \sqrt{t} \|\mathcal{H}^{(l)}\|_2 + t \|\mathcal{H}^{(l)}\|_\infty) 2 \exp(-t) \mathrm{d} t
\end{align*}
where we define $\|\mathcal{H}^{(l)}\|_{*} := \sup_{G \in \mathcal{G}^{(T-1)} }\|G\|_{*}$ for $*=2$ and $\infty$.
This leads to the right hand bound as 
$
C \sqrt{\mathrm{Tr}[\tilde{A}^{2k}]} \left( \mathbb{E}_{\epsilon'}\left[ \sup_{G \in \mathcal{G}^{(T-1)} } \left| \epsilon'^\top G \right| \right] + \|\mathcal{H}^{(l)}\|_2 \right)
$
for a universal constant $C$, where we used $2p \leq 1$.
using the assumption that the output is bounded by the activation function, we take $\sup_{Z \in \mathcal{H}^{(l)}}\|Z\|_2 \leq \sqrt{N}R =: D$.
Now, we have
\begin{align}
\Pi &\leq \bigg[C \sqrt{\mathrm{Tr}[\tilde{A}^{2k}]}  \bigg( \mathbb{E}_{\boldsymbol{\epsilon}}\sup_{G \in \mathcal{G}^{(T-1)}}  \bigg\vert \sum_{n=1}^{N} \epsilon_{n} G_{n} \bigg\vert + D \bigg) \bigg] \nonumber\\
&\leq \bigg[C \sqrt{\bigg(\sum_{i=1}^{N}\lambda_{i}^{k}\bigg)^{2} }  \bigg( \mathbb{E}_{\boldsymbol{\epsilon}}\sup_{G \in \mathcal{G}^{(T-1)}}  \bigg\vert \sum_{n=1}^{N} \epsilon_{n} G_{n} \bigg\vert + D \bigg) \bigg] \nonumber\\
&\leq \bigg[C \sum_{i=1}^{N} \vert\lambda_{i}\vert^{k} \bigg( \mathbb{E}_{\boldsymbol{\epsilon}}\sup_{G \in \mathcal{G}^{(T-1)}}  \bigg\vert \sum_{n=1}^{N} \epsilon_{n} G_{n} \bigg\vert + D \bigg) \bigg]. \label{eq:simplyfied_Pi_1}
\end{align}

Further, we can bound the last term of \eqref{eq:reduction_rad-step2} by
\begin{equation*} 
\begin{split}
R.H.S  &= (1-\mu)B^{(T+1)}B^{(A)}\mathbb{E}_{\boldsymbol{\epsilon}}  \bigg[ \sup_{   } \bigg\vert  \sum_{n=1}^{N} \epsilon_{n}  \bar{A}_{n} \bigg\vert \bigg]\\
& \leq 2 (1-\mu)B^{(T+1)}B^{(A)}\mathbb{E}_{\boldsymbol{\epsilon}} \Bigg[\Bigg\| \sum_{n=1}^{N} \epsilon_{n} \bar{A}_{n\cdot}   \Bigg\|_{2} \Bigg] \\
& \leq 2 (1-\mu)B^{(T+1)}B^{(A)}\ \sqrt{ \mathbb{E}_{\boldsymbol{\epsilon}} \sum_{c=1}^{C_0} \Bigg( \sum_{n=1}^{N} \epsilon_{n} \bar{A}_{nc}  \Bigg)^{2} }  \;\; \mathrm{(Jensen\;Inequality)}\\
& = 2(1-\mu)B^{(T+1)}B^{(A)}\sqrt{ \mathbb{E}_{\boldsymbol{\epsilon}} \sum_{c=1}^{C_0} \sum_{n,m=1}^{N}  \epsilon_{n}\epsilon_{m} \bar{A}_{nc}\bar{A}_{mc}   } \\
& = 2(1-\mu)B^{(T+1)}B^{(A)}\sqrt{  \sum_{c=1}^{C_0} \sum_{m=1}^{N} 2p (\bar{A}_{mc})^{2}   } \\
& = 2 (1-\mu)B^{(T+1)}B^{(A)}\sqrt{  2p   } \| \bar{A} \|_{\mathrm{F}}\\
& \leq 2 (1-\mu)B^{(T+1)}B^{(A)}\sqrt{  2p   } \sum_{i=1}^{N} |\lambda_{i}|.
\end{split} 
\end{equation*} 

Uisng the assumption that $p = p_0 = \frac{MU}{(M+U)^2}$, we have
\begin{equation} 
R.H.S   \leq 2^{5/2} (1-\mu)\frac{B^{(T+1)}B^{(A)}\sqrt{MU}}{(M+U)} \sum_{i=1}^{N} |\lambda_{i}|.
\label{eq:simplyfied_LINK_componenet}
\end{equation}

Now applying \eqref{eq:simplyfied_Pi_1} and \eqref{eq:simplyfied_LINK_componenet} to \eqref{eq:reduction_rad-step2} for $p = p_{0}$, we have
\begin{equation} 
\begin{split}
Q^{-1}\mathcal{\bar{R}}(\mathcal{\tilde{F}_{\mu,\gamma}}, p_{0})  
&C\leq B^{(T+1)}\mu\bigg[\bigg(I + \sum_{k=1}^{L-1} B^{(T)k} K \gamma^{k} \sum_{j=1}^{N} |\lambda_j|^k   +  \gamma^{L} B^{(T)L} \sum_{j=1}^{N} |\lambda_j|^L \bigg)  \mathbb{E}_{\boldsymbol{\epsilon}}  \bigg\vert  \sup_{G \in \mathcal{G}^{(T)} }  \sum_{n=1}^{N} \epsilon_{n} G_{n} \bigg\vert   \\
& \qquad\qquad\qquad + \bigg(\sum_{k=1}^{L-1} B^{(T)k} K \gamma^{k} \sum_{j=1}^{N} |\lambda_j|^k   +  \gamma^{L} B^{(T)L} \sum_{j=1}^{N} |\lambda_j|^L \bigg)D\bigg]\\   
&\qquad\qquad\qquad\qquad\qquad\qquad\qquad + (1-\mu)2^{5/2} \frac{B^{(T+1)}B^{(A)}\sqrt{MU}}{(M+U)} \sum_{i=1}^{N} |\lambda_{i}| 
\end{split} \label{eq:reduction_rad-step1}
\end{equation}

Considering the node feature based layers in \eqref{eq:reduction_rad-step1}, we can apply simplification as
\begin{align*} 
\mathcal{\bar{R}}(\mathcal{\tilde{G}}^{(T)},p)  & 
\leq 2 \mathcal{\bar{R}}(\mathcal{G}^{(T)},p) \nonumber \\
& = 2\mathbb{E}_{\boldsymbol{\epsilon}} \Bigg[\sup_{G_{\cdot c} \in\mathcal{G}^{(T-1)} , w \in \mathbb{R}^{C_T}:\|w\|_1 \leq B^{(T)}} 
\Bigg\vert  \sum_{n=1}^{N}  \epsilon_{n} \sum_{c=1}^{C_T} G_{nc}  w_c \Bigg\vert \Bigg] \nonumber \\
& = 2\mathbb{E}_{\boldsymbol{\epsilon}} \Bigg[\sup_{Z_{\cdot c} \in\mathcal{G}^{(T-1)} ,  w \in \mathbb{R}^{C_T}:\|w\|_1 \leq B^{(T)}} 
\Bigg\vert \sum_{c=1}^{C_T} \sum_{n=1}^{N}  \epsilon_{n}  G_{nc}  w_c \Bigg\vert \Bigg] \nonumber \\
& = 2B^{(T)}\mathbb{E}_{\boldsymbol{\epsilon}} \Bigg[\sup_{G_{n \cdot} \in\mathcal{G}^{(T-1)} } 
\Bigg\vert  \sum_{n=1}^{N}  \epsilon_{n}  G_{n \cdot}   \Bigg\vert \Bigg], 
\end{align*}
which by repeating on all hypothesis classes $\mathcal{G}^{(l)},\;l=T-1,\ldots,1$, we obtain
\begin{align} 
\mathcal{\bar{R}}(\mathcal{\tilde{G}}^{(1)},p)  & = 2^{T-1} \prod_{l=1}^{T-1}B^{(l)}\mathbb{E}_{\boldsymbol{\epsilon}} \Bigg[\sup_{G_{n \cdot} \in\mathcal{G}^{(0)} } 
\Bigg\vert  \sum_{n=1}^{N}  \epsilon_{n}  G_{n \cdot}   \Bigg\vert \Bigg].  \label{eq:bound_H_L} 
\end{align}

Now, we bound 
\begin{align*} 
\mathcal{\bar{R}}(\mathcal{\tilde{G}}^{(0)},p)  & 
\leq 2 \mathcal{\bar{R}}(\mathcal{G}^{(0)},p)  \nonumber \\
& = 2\mathbb{E}_{\boldsymbol{\epsilon}} \Bigg[\sup_{w \in \mathbb{R}^{C_0}:\|w\|_1 \leq B^{(0)}} 
\Bigg\vert  \sum_{n=1}^{N} \sum_{c=1}^{C_0} \epsilon_{n} X_{nc}  w_c \Bigg\vert \Bigg]  \nonumber  \\
 & = 2 B^{(0)} \mathbb{E}_{\boldsymbol{\epsilon}} \Bigg[\max_{c \in [C_0]} \Bigg\vert  \sum_{n=1}^{N} \epsilon_{n} X_{nc} \Bigg\vert \Bigg]  \nonumber \\
& \leq 2 B^{(0)}\mathbb{E}_{\boldsymbol{\epsilon}} \Bigg[\Bigg\| \sum_{n=1}^{N} \epsilon_{n} X_{n\cdot}   \Bigg\|_{2} \Bigg]  \nonumber \\
& \leq 2 B^{(0)} \sqrt{ \mathbb{E}_{\boldsymbol{\epsilon}} \sum_{c=1}^{C_0} \Bigg( \sum_{n=1}^{N} \epsilon_{n} X_{nc}  \Bigg)^{2} }  \;\; \mathrm{(Jensen\;Inequality)} \nonumber \\
& = 2B^{(0)}\sqrt{ \mathbb{E}_{\boldsymbol{\epsilon}} \sum_{c=1}^{C_0} \sum_{n,m=1}^{N}  \epsilon_{n}\epsilon_{m} X_{nc}X_{mc}   }  \nonumber  \\
& = 2B^{(0)}\sqrt{  \sum_{c=1}^{C_0} \sum_{m=1}^{N} 2p (X_{mc})^{2}   } \nonumber \\
& = 2 B^{(0)}\sqrt{  2p   } \| X \|_{\mathrm{F}}. 
\end{align*}

Furthermore, given that $p = p_0 = \frac{MU}{(M+U)^2}$, we have
\begin{align}
\mathcal{\bar{R}}(\mathcal{\tilde{G}}^{(0)},p_0) \leq  2 B^{(0)}\sqrt{\frac{2MU}{(M+U)^2}}\| X \|_{\mathrm{F}}. \label{eq:rad_h0}
\end{align}
and combining \eqref{eq:bound_H_L}  and \eqref{eq:rad_h0} with \eqref{eq:reduction_rad-step1}, 
we arrive at the final bound
\begin{equation*} 
\begin{split}
Q^{-1}\mathcal{\bar{R}}(\mathcal{\tilde{F}_{\mu,\gamma}}, p_{0})  
&\leq C'B^{(T+1)}\mu\bigg[ 2^{T}\prod_{l=0}^{T-1}B^{(l)} \sqrt{\frac{2MU}{(M+U)^2}}\bigg(I + \sum_{k=1}^{L-1} B^{(T)k} L \gamma^{k} \sum_{j=1}^{N} |\lambda_j|^k  \\
&\quad +  \gamma^{K} B^{(T)L} \sum_{j=1}^{N} |\lambda_j|^L \bigg) \| X \|_{\mathrm{F}}   + \bigg(\sum_{k=1}^{L-1} B^{(T)k} L \gamma^{k} \sum_{j=1}^{N} |\lambda_j|^k \\
&\quad +  \gamma^{K} B^{(T)L} \sum_{j=1}^{N} |\lambda_j|^L \bigg)D\bigg] + (1-\mu)2^{5/2} \frac{B^{(T+1)}B^{(A)}\sqrt{MU}}{(M+U)} \sum_{i=1}^{N} |\lambda_{i}|.  
\end{split} 
\end{equation*}
}

\textit{Proof of Theorem 2.}{
We modify the following hypothesis class \eqref{eq:hypo_GPCN-LINK} to suit AGPCN-LINK as
\begin{multline*}
\mathcal{F}_{\mu,\gamma} = \Big\{ X,\bar{A} \mapsto 
\mathrm{softmax} \bigg(f^{(3)} \circ (\mu f^{(1)} \circ g^{(T)} \circ \cdots \circ g^{(0)}(X) + (1 - \mu)  f^{(2)}(\bar{A})) \vert \\
 g^{(l)}(Z) = \mathrm{Relu}(ZW^{(l)}),  f^{(1)}(X_T) = \theta_{0}X_{T} + \sum_{k=1}^{L} \theta_{k}\bar{A}^{k} X_{T} W^{(T)k}  \bigg), \\  
f^{(2)}(\bar{A})) = \bar{A}W^{(A)},\; f^{(3)}(Z) = ZW^{(T+1)},\; \|W_{ \cdot c}^{(l)} \|_{1} \leq  B^{(l)}\; 
 \mathrm{for\; all} \; c \in [C_{l+1}], \|W_{ \cdot c}^{(A)} \|_{1} \leq  B^{(A)}
\Big\}. 
\end{multline*}

Following a similar proof procedure as in Theorem 1 we arrive at the bound the desired bound. 
}

\section{Experiments with Homophilous Data}
Table \ref{table:tb_homophilic} shows performances for homophilous graphs Cora, Citeseer, and Pubmed from \cite{Pei-20-Geom-GCN}. Both GPCN and GPCN-LINK  have obtained with low accuracy for Cora and Citeseer while GCNII, GPRGNN, and APPNP have given the best accuracy. Further, note that LINKX only gives a weak accuracy for these datasets. For the Pubmed dataset, GPCN and GPCN-LINK have given the best accuracy outperforming previous best results from GCNII, GPRGNN, and APPNP. Overall, GPCN and its varaints have shown the ability obtain better accuracy compared to LINK and LINKX while giving competitive performances with respect to other graph convolution models.

\begin{table}[t]
\center
\begin{tabular}{llll}
Method & Cora  &  Citeseer & Pubmed      \\ \hline
Classes & 7 &   4 & 3  \\
Nodes & 2708 &   3327 & 19717  \\
Edges & 5429 &   4732 & 44338  \\
Features & 1433 &   3703 & 500 \\
Edge Homophily & 0.825 &   0.718 & 0.792  \\
\hline
MLP   & 74.46$\pm$2.08 & 73.41$\pm$1.82 &  87.60$\pm$0.31  \\
GCN & 87.22$\pm$1.09 &  76.26$\pm$1.49 & 88.11$\pm$0.45   \\
SGC & 87.22$\pm$1.06 &   76.37$\pm$1.49 & 88.01$\pm$0.49 \\
GCNII & \textbf{88.16}$\pm$\textbf{1.20} &  \textbf{76.95}$\pm$\textbf{1.48} & 89.48$\pm$0.59 \\
GPRGNN & \textbf{87.76}$\pm$\textbf{1.25} &  \textbf{76.80}$\pm$\textbf{1.59} & 89.43$\pm$0.57  \\
APPNP  & \textbf{88.00}$\pm$\textbf{1.16} &  \textbf{77.19}$\pm$\textbf{1.86} &  89.38$\pm$0.38  \\
LINK   & 80.88$\pm$1.35 & 65.41$\pm$3.49 &  81.16$\pm$0.32  \\
LINKX   & 84.64$\pm$1.13 & 73.19$\pm$0.99 &  87.86$\pm$0.77  \\
GPCN   & 86.29$\pm$1.33 & 76.16$\pm$1.76 &  \textbf{89.72}$\pm$\textbf{0.49}  \\
GPCN-LINK   & 86.03$\pm$0.85 & 75.45$\pm$1.64 &  \textbf{89.85}$\pm$\textbf{0.33} \\
AGPCN   & 86.23$\pm$1.65 & 74.86$\pm$1.52 &  \textbf{89.48}$\pm$\textbf{0.50}  \\
AGPCN-LINK   & 85.25$\pm$2.73 & 74.08$\pm$3.71 &  89.21$\pm$0.45  \\
\end{tabular}
\caption{Properties and node-classification accuracy of homophilic datasets. The best three results are highlighted. }\label{table:tb_homophilic}
\end{table}

\section{Ablation Studies}
We carried out ablation studies to understand the robustness of proposed models against oversmoothing due to polynomial structure of scaling parameters and weights. For ablation experiments, we select the best hyperparameter selection for a dataset and change the scaling parameter ($\gamma$) in the range of $2^{0},2^{-2},\ldots,2^{-8}$, residual layers $L=1,2,4\ldots,16$, and dropout $\in (0.0,0.3,0.6,0.9)$ while keeping the rest of the parameters (learning rate, hidden,...etc) fixed. 

\begin{figure}
\centering
\begin{minipage}{.5\textwidth}
  \centering
 \includegraphics[width=.99\linewidth]{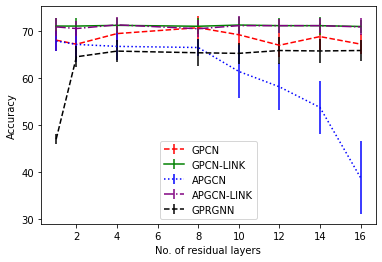}
  \captionof{figure}{Ablation study on Chameleon}
  \label{fig:cham}
\end{minipage}%
\begin{minipage}{.5\textwidth}
  \centering
  \includegraphics[width=.99\linewidth]{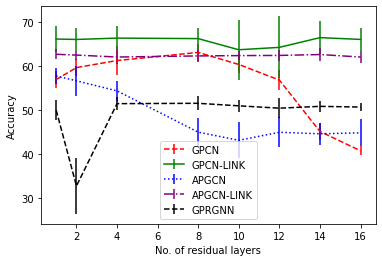}
  \captionof{figure}{Ablation study on Squirrel}
  \label{fig:squirrel}
\end{minipage}
\end{figure}

Figure \ref{fig:cham} shows that node calssification accuracy for Chameleon with GPCN-LINK and APGCN-LINK do not decrease in accuracy as the number layers increases, hence, rubust against oversmoothing. GPCN is relatively stable with increasing layers though the accuracy is low compared to  GPCN-LINK and APGCN-LINK. APGCN shows strong oversmoothing, however, gives a higher accuracy compared to GPRGNN with one residual layer. Similar behaviours are seen with Squirrel (Figure \ref{fig:squirrel}) where GPCN and AGPCN show oversmoothing and other models are robust against oversmoothing. 

A common feature among all the proposed methods is that they provide higher accuracy at low number of residual layers. This observation is with agreement with the theoretical analysis in Section 4.  

\section{Hyperparameter Summary}
In this section, we provide hyperparameters selected for out proposed methods. Hyperparameters  are selected for all models (where applicable) from  hidden $\in \{64,512\}$, learning parameter $\in \{0.01,0.05\}$ for the Adam method, weight decay $\in \{0.0, 0.001,0.00001\}$, initial feature learning layers $T \in \{1,2,3,4,5\}$, residual layers $L \in \{1,2,4,8\}$, $\gamma \in \{2^{8},2^{6},\ldots,2^{0},2^{-2},\ldots,2^{6}\}$, dropout $\in \{0,0.3,0.6,0.9\}$.

\begin{table}[t]
\begin{tabular}{l|l|l|l|l|l|l|l}
Dataset & Learning rate & Hidden & Weight decay & T & MLP & $\gamma$  & Dropout  \\ \hline
Cora & 0.05 & 64 & 0.001 & 2 & 1 & 1 & 0.6   \\
Citeseer & 0.01 & 64 & 0.001 & 2 & 1 & 0.25  & 0.3 \\
Pubmed & 0.01 & 64 & 0.001 & 4 & 2 &  0.0625  & 0.3 \\
Chameleon & 0.01 & 512 & 0.001 & 8 & 1 & 0.25  & 0.3 \\
Squirrel & 0.05 &  512& 1e-5 & 8 & 1   & 0.0625  & 0.3 \\
Actor & 0.05 & 512 & 0.01 & 2 & 2 & 0.0625   &  0.0 \\
Cornell & 0.01 & 512 & 0.001 & 2 & 4   &  & 0.3 \\
Texas & 0.01 & 512 & 0.001 & 1 & 3 &  0.015625  & 0.6 \\
Wisconsin  & 0.05 & 512 & 0.001 & 2   & 1 & 0.0625 & 0.3 \\
Twitch-DE   & 0.01 & 512 & 0.001 & 2  & 1 & 16 & 0.0 \\
Penn94      & 0.05  & 512 & 1e-5 & 4 & 1 &  0.0625 & 0.0 \\
Yelp-Chi     & 0.01 & 64 & 1e-5 & 4 & 4 & 0.0625 & 0.0\\
Deezer-Europe      & 0.01 & 64 & 1e-5 & 16 & 5 & 0.0625 & 0.0\\
Genius    & 0.01 & 64 & 1e-5 & 8 & 1 & 0.0625 & 0.0\\
\end{tabular}
\caption{Hyperparameters for GPCN}
\end{table}

\begin{table}[t]
\begin{tabular}{l|l|l|l|l|l|l|l|l}
Dataset & Learning rate & Hidden & Weight decay & T & MLP & $\gamma$ &   Dropout  \\ \hline
Cora &  0.01 & 512 & 0.001 & 2 & 1 & 4 & 0.6    \\
Citeseer & 0.001 & 64 & 1e-5 & 1 & 1 & 64 & 0.3   \\
Pubmed & 0.01 & 512 & 0.001 & 4 & 3 & 0.0625 &  0.3  \\
Chameleon & 0.01  & 512 & 0.001 & 4 & 2 & 0.00390625 & 0.6   \\
Squirrel & 0.05 & 512 & 1e-5 & 8 & 1 & 4 &   0 \\
Actor & 0.01 & 64 & 0.001 & 4 & 3 & 0.0625  & 0.0   \\
Cornell & 0.01 & 512 & 0.001 & 1 & 4 & 0.015625 & 0  \\
Texas & 0.01 & 512 & 0.001 & 1 & 4 & 0.00390625 & 0  \\
Wisconsin  & 0.01 & 512 & 0.001 & 2 & 3 & 0.25 & 0  \\
Twitch-DE  &  0.01 & 64 & 0.001 & 2 & 2 & 64 & 0.0   \\
Penn94  &  0.01 & 512 & 1e-5 & 4 & 1 & 4 & 0.0  \\
Yelp-Chi  & 0.01 & 512 & 1e-5 & 2 & 5 & 0.015625 & 0.0 \\
Deezer-Europe  & 0.01 & 512 & 0.001 & 1 & 8 &  3  & 0.0\\
Genius  & 0.01 & 64 & 0.001 & 8 & 2 & 1 & 0.0  \\
\end{tabular}
\caption{Hyperparameters for GPCN-LINK}
\end{table}

\begin{table}[t]
\begin{tabular}{l|l|l|l|l|l|l}
Dataset & Learning rate & Hidden & Weight decay & T & MLP   & Dropout  \\ \hline
Cora & 0.05 & 512 & 0.001 & 2 &  1  & 0.6 \\
Citeseer & 0.01 & 512 & 0.001 & 2   & 1 & 0.6 \\
Pubmed & 0.01 & 64  & 0.0 & 2 &  2  & 0.3 \\
Chameleon & 0.01 & 512 & 0.001 & 1   & 1 & 0.6 \\
Squirrel & 0.01  & 64 & 0.001 & 1   & 1 & 0.3  \\
Actor & 0.05 & 64 & 0.001 & 2 & 1   & 0.3 \\
cornell & 0.05 & 512 & 0.001   & 1 & 2 & 0.3 \\
Texas & 0.05 & 512 & 0.001 & 1   & 2 & 0.3 \\
Wisconsin  & 0.05 & 512 & 0.001 & 1   & 2  & 0.3 \\
Twitch-DE   & 0.01 & 512 & 1e-5 & 2   & 1  & 0.0 \\
Penn94    & 0.01 & 64 & 1e-5 & 8 & 2   & 0.3\\
Yelp-Chi     & 0.01 & 512 & 1e-05 & 2   & 2 & 0.0\\
Deezer-Europe   & 0.01 & 64 & 1e-5 & 1 & 2 & 0.6\\
Genius    & 0.01 & 512 & 1e-5 & 2 & 2 & 0.0 \\
\end{tabular}
\caption{Hyperparameters for AGPCN}
\end{table}

\begin{table}[t]
\begin{tabular}{l|l|l|l|l|l|l}
Dataset & Learning rate & Hidden & Weight decay & T & MLP   & Dropout  \\ \hline
Cora & 0.01 & 512 & 0.001 & 2 & 1   & 0.6 \\
Citeseer & 0.01 & 512 & 0.001 & 2   & 1 & 0.6 \\
Pubmed & 0.01 & 512 & 0.001 & 2 & 2   & 0.3 \\
Chameleon & 0.05 & 64 & 0.001  &  2  & 1 & 0.3 \\
Squirrel & 0.05 & 512 & 1e-5 &  8  & 1 & 0.3 \\
Actor & 0.01  & 512 & 0.001 & 8 & 2   & 0.0 \\
cornell & 0.01 & 64 & 0   & 4 & 3 & 0.0\\
Texas &  0.01 & 512 & 0.001 & 1   & 4 & 0.3\\
Wisconsin  & 0.01 & 512 & 0.001 & 2   & 2 & 0.0\\
Twitch-DE   &  0.01 & 64 & 0.001 & 4   & 2 & 0 \\
Penn94    & 0.01 & 64 & 1e-5 & 4 & 2   & 0.3\\
Yelp-Chi     & 0.01 &  512& 1e-5 &  2  & 3 & 0.0\\
Deezer-Europe  & 0.01 & 512   & 1e-5 & 1 & 2 & 0.6\\
Genius    & 0.01 & 64 & 1e-5 & 4 & 2 & 0.3 \\
\end{tabular}
\caption{Hyperparameters for AGPCN-LINK}
\end{table}

\end{document}